\documentclass{Interspeech}

\usepackage{multirow}

\interspeechcameraready

\title{Assessing the feasibility of Large Language Models for detecting micro-behaviors in team interactions during space missions}

\author[affiliation={1}]{Ankush}{Raut}
\author[affiliation={1}]{Projna}{Paromita}
\author[affiliation={2}]{Sydney}{Begerowski}
\author[affiliation={2}]{Suzanne}{Bell}
\author[affiliation={1}]{Theodora}{Chaspari}

\affiliation{}{University of Colorado Boulder}{USA}
\affiliation{}{National Aeronautics and Space Administration}{USA}
\email{\{ankush.raut,projna.paromita,theodora.chaspari\}@colorado.edu, \{sydney.r.begerowski,suzanne.t.bell\}@nasa.gov}
\keywords{large language models, zero-shot classification, in-context learning, quantization, micro-behaviors, teams}

\usepackage{comment}

\begin{document}

\maketitle

\begin{abstract}\vspace{-3pt}
   We explore the feasibility of large language models (LLMs) in detecting subtle expressions of micro-behaviors in team conversations using transcripts collected during simulated space missions. Specifically, we examine zero-shot classification, fine-tuning, and paraphrase-augmented fine-tuning with encoder-only sequence classification LLMs, as well as few-shot text generation with decoder-only causal language modeling LLMs, to predict the micro-behavior associated with each conversational turn (i.e., dialogue). Our findings indicate that encoder-only LLMs, such as RoBERTa and DistilBERT, struggled to detect underrepresented micro-behaviors, particularly discouraging speech, even with weighted fine-tuning. In contrast, the instruction fine-tuned version of Llama-3.1, a decoder-only LLM, demonstrated superior performance, with the best models achieving macro F1-scores of 44\% for 3-way classification and 68\% for binary classification. These results have implications for the development of speech technologies aimed at analyzing team communication dynamics and enhancing training interventions in high-stakes environments such as space missions, particularly in scenarios where text is the only accessible data.
\end{abstract}

\vspace{-3pt}
\section{Introduction}
\vspace{-5pt}
Given their effectiveness in spoken language understanding (SLU)~\cite{mousavi2024llms,zhu2024zero,he2023can}, competence as few-shot learners~\cite{brown2020languagemodelsfewshotlearners}, and reasoning abilities~\cite{wei2022chainofthoughtpromptingelicitsreasoning}, large language models (LLMs) can potentially infer obscure meanings and semantics from speech in interpersonal communication settings, such as dyadic or multiparty interactions. This capability enables their application across diverse domains, including mental health assessment, educational tutoring systems, and workplace collaboration analysis, where labeled data is sparse and understanding conversational context and intent is crucial for providing meaningful feedback and support. In teamwork settings, LLMs can serve as support tools to foster team cohesion by enhancing informal knowledge-sharing processes~\cite{callari2025can} and improving work efficiency through personalized response generation in professional interactions~\cite{bastola2023llm}. Additionally, LLM-powered chatbots have the potential to act as virtual liaisons, facilitating seamless communication, personalized support, and conflict resolution in workplace teamwork~\cite{oluwagbade2024conversational}.

Micro-behaviors, defined as momentary, subtle linguistic and paralinguistic indicators of thoughts and feelings toward another team member~\cite{cortina2013selective,smith2022microaggressions} can offer unique insights into operationally-relevant team functioning. Depending on how they are communicated, micro-behaviors can uplift or discourage a team member~\cite{begerowski2023s}. Repeated micro-behaviors can also have lasting effects on team interactions. Identifying these momentary acts within team interactions can enhance our understanding of factors influencing team states and performance. Given their transient nature, manually detecting micro-behaviors is highly time-consuming. Therefore, developing automated models for their detection could advance speech technologies that support teamwork.

Here, we examine the feasibility of using LLMs to detect micro-behaviors in team interactions during simulated space missions. We investigate the zero-shot micro-behavior classification ability and the fine-tuning of encoder-only sequence classification LLMs, including the Robustly Optimized Bidirectional Encoder Representations from Transformers (BERT) Pretraining Approach (RoBERTa) and DistilBERT, and compare them with few-shot prompting in the instruction fine-tuned version of a decoder-only causal language modeling LLM, including Llama-3.1. Our analysis includes a 3-way classification task, categorizing each turn as uplifting, discouraging, or neither, as well as a binary classification task distinguishing between uplifting and discouraging turns. Results indicate that LLMs are feasible for this task, with decoder-only causal LLMs outperforming encoder-only models, particularly in detecting underrepresented classes with a significantly higher recall.


\vspace{-3pt}
\section{Prior Work}
\vspace{-5pt}
Recent research has explored automated methods for detecting micro-aggressions, a term used for commonplace verbal, behavioral, or environmental slights that communicate hostile, derogatory, or negative attitudes toward members of marginalized groups~\cite{sue2010microaggressions}. These methods include handcrafted lexicons~\cite{ali2020automated}, unigrams/bi-grams, and topic models~\cite{breitfeller2019finding}. Recent studies have also investigated use of LLMs for the same task. Caselli et al.'s study on unsupervised microaggression detection computed the embeddings for a set of manually curated seed words known to exhibit gender and racial bias, then compared those seed words against target tokens in the corresponding embedding space~\cite{sabri2021leveraging}. Embeddings from encoder-only LLMs such as BERT and RoBERTa have also been evaluated for their ability to detect microaggressions in written text~\cite{tareque2024overview} and scripted conversational speech~\cite{ngueajio2023towards}.

The contributions of this paper compared to prior work are as follows: (1) Beyond micro-aggressions, this study aims to detect both positive (uplifting) and negative (discouraging) micro-behaviors--an area that remains largely unexplored~\cite{cortina2013selective}. Unlike micro-aggressions, which often carry gendered or racial connotations, micro-behaviors can be either positive or negative and are highly context-dependent, making their detection more nuanced. (2) The majority of prior work has focused on written speech rather than oral multiparty communication which is contextually more complex and semantically richer. (3) Prior work on LLMs has examined encoder-only sequence classification with pre-trained word embeddings~\cite{sabri2021leveraging,tareque2024overview,ngueajio2023towards}, while decoder-only models that leverage autoregressive generation and in-context learning have not been explored despite their potential advantages in handling nuanced language and contextual dependencies. Additionally, our work focuses on micro-behavior detection using only text transcripts of conversations, making this task even more challenging, particularly in the absence of audio data, which enhances models' ability to detect signals corresponding to the expression of micro-behaviors.

\vspace{-3pt}
\section{Data Description}
\label{sec:Data}
\vspace{-5pt}
Our data came from five teams that participated in a 45-day mission at the Human Exploration Research Analog (HERA) of the U.S. National Aeronautics and Space Administration (NASA). Each team took part in a simulated space expedition, performing a geological exploration scenario of the journey to and from Mars’s moon, Phobos. In this paper, we used data collected from the team interaction battery (TIB) tasks that were performed five times (one pre-mission, four in-mission) per crew. Each TIB task had an average duration of 1.5 hours and included a 45-minute decision-making task, in which teams could only arrive at the correct decision through adequate information sharing and teamwork, and a relational task in which crew members responded to various questions that foster interpersonal relations among crew. Depending on the task, each conversational turn (i.e., dialogue) was characterized into an event type as task-related, relational, or a transition period event (i.e., in-between the two tasks). Micro-behaviors were coded according to an adapted version of Smith \& Griffins's (2022) theoretical framework in terms of Violation (i.e., presence of valenced behavior; uplifting/positive or discouraging/negative), Intensity (i.e., force of behavior in terms of how uplifting or discouraging is the behavior), and Intent (i.e., motive of the behavior in terms of whether it was deliberate or unintentional). This paper examines only the Violation dimension in 13,058 conversational turns: 17.8\% uplifting, 3.3\% discouraging, 75.76\% neither, and 3.14\% unlabeled.

\begin{figure}[t]
    \centering    \includegraphics[width=0.9\linewidth]{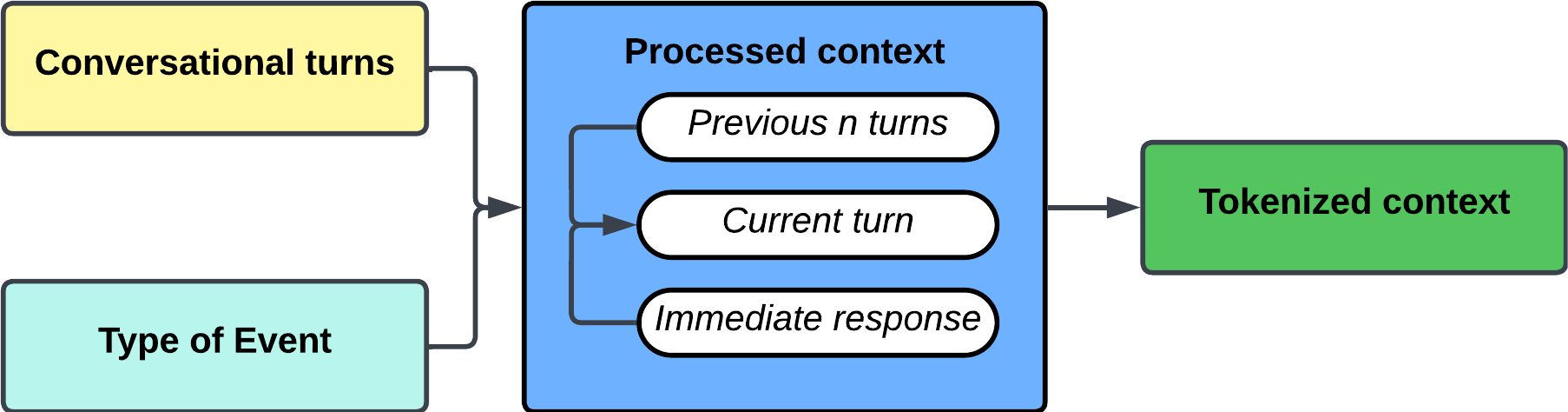}
    \caption{Illustration of the pre-processing steps for encoder-only sequence classification.}
    \label{fig:encoderonly}
\end{figure}

\begin{figure}[t]
    \centering
    \includegraphics[width=1\linewidth]{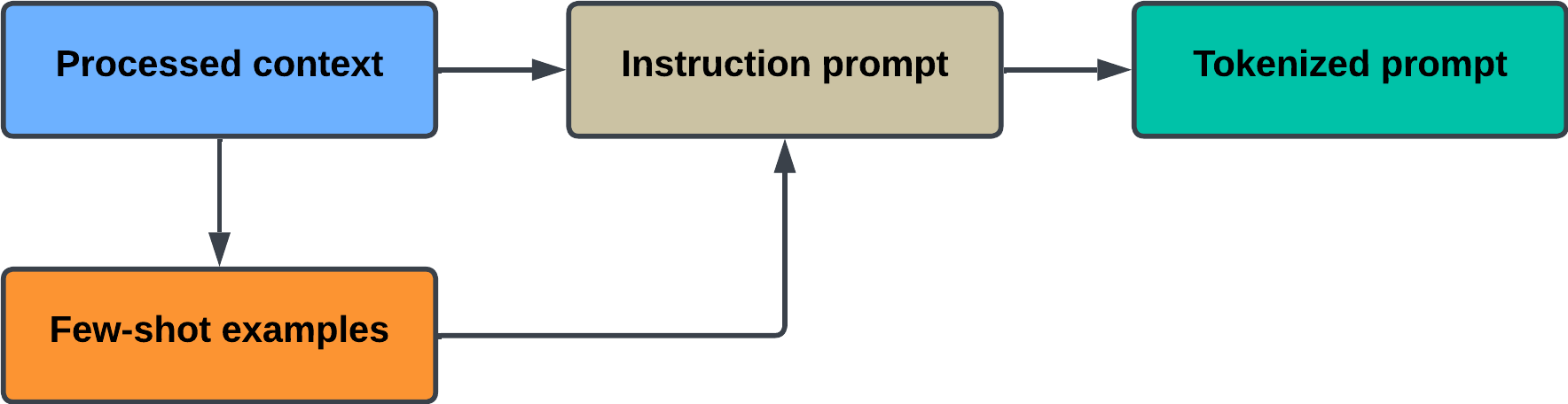}
    \caption{Illustration of the instruction prompt creation pipeline for decoder-only causal text generation.}
    \label{fig:decoderonly}
\end{figure}

\vspace{-3pt}
\section{Methodology}
\vspace{-5pt}
In this section, we describe the encoder-only sequence classification (Section~\ref{subsec:EncoderOnly}) and the decoder-only causal text generation (Section~\ref{subsec:DecoderOnly}) methods which are used in the 3-way (i.e., uplifting, discouraging, and neither) and binary (i.e., uplifting and discouraging) micro-behavior classification tasks.

\subsection{Encoder-only Sequence Classification}
\vspace{-5pt}
\label{subsec:EncoderOnly}
The encoder-only sequence classification pipeline takes the conversational turns along with additional task-related context, tokenizes it, and processes it through encoder-only LLMs to create a meaningful representation, which is then used to predict the micro-behavior label for each conversational turn (Figure \ref{fig:encoderonly}).
The input to the model includes the current conversational turn in which the micro-behavior label needs to be determined, its event type (Section~\ref{sec:Data}), the $k$ previous conversational turns, and the response to the current turn. We experiment with including $k=3,4$ previous turns, presenting the results for both configurations separately. Adding over 4 previous turns leads to diminishing returns and, in some scenarios, performance degradation, which is why we do not report results from experiments involving over 4 turns.  We don't report the results when adding only 2 previous turns either, as it is too small a context for the models to appropriately interpret the conversation, confirmed by the poor results on this configuration. This complete context is then tokenized and passed to the sequence classification LLMs.

We experiment with the state-of-the-art sequence classification LLMs RoBERTa~\cite{liu2019robertarobustlyoptimizedbert} and DistilBERT~\cite{sanh2020distilbertdistilledversionbert}. We use the twitter-roberta-base-sentiment-latest model~\cite{twitter_roberta}, a RoBERTa-base model fine-tuned for sentiment analysis with the TweetEval benchmark by CardiffNLP, for 3-way classification. This model returns negative, neutral, and positive labels, which we map to discouraging, neither, and uplifting micro-behavior labels, respectively. During tokenization, the input context is left-truncated to 512 tokens, which is the maximum sequence length for RoBERTa. 

We use the distilbert-base-uncased-finetuned-sst-2-english model~\cite{distilbert_sst2}, a DistilBERT-base-uncased model fine-tuned on the Stanford Sentiment Treebank (SST2) dataset, for binary classification on conversational turns that are valenced, i.e., labeled as either discouraging or uplifting. This model returns negative and positive sentiment, which we map to discouraging and uplifting micro-behaviors, respectively. The goal of experimenting with this configuration is to evaluate the ability of a BERT-like LLM in distinguishing discouraging micro-behaviors from uplifting ones. The tokenization specifications for this model are the same as that used for RoBERTa.


\begin{figure*}[t]
    \centering
    \includegraphics[width=0.8\linewidth]{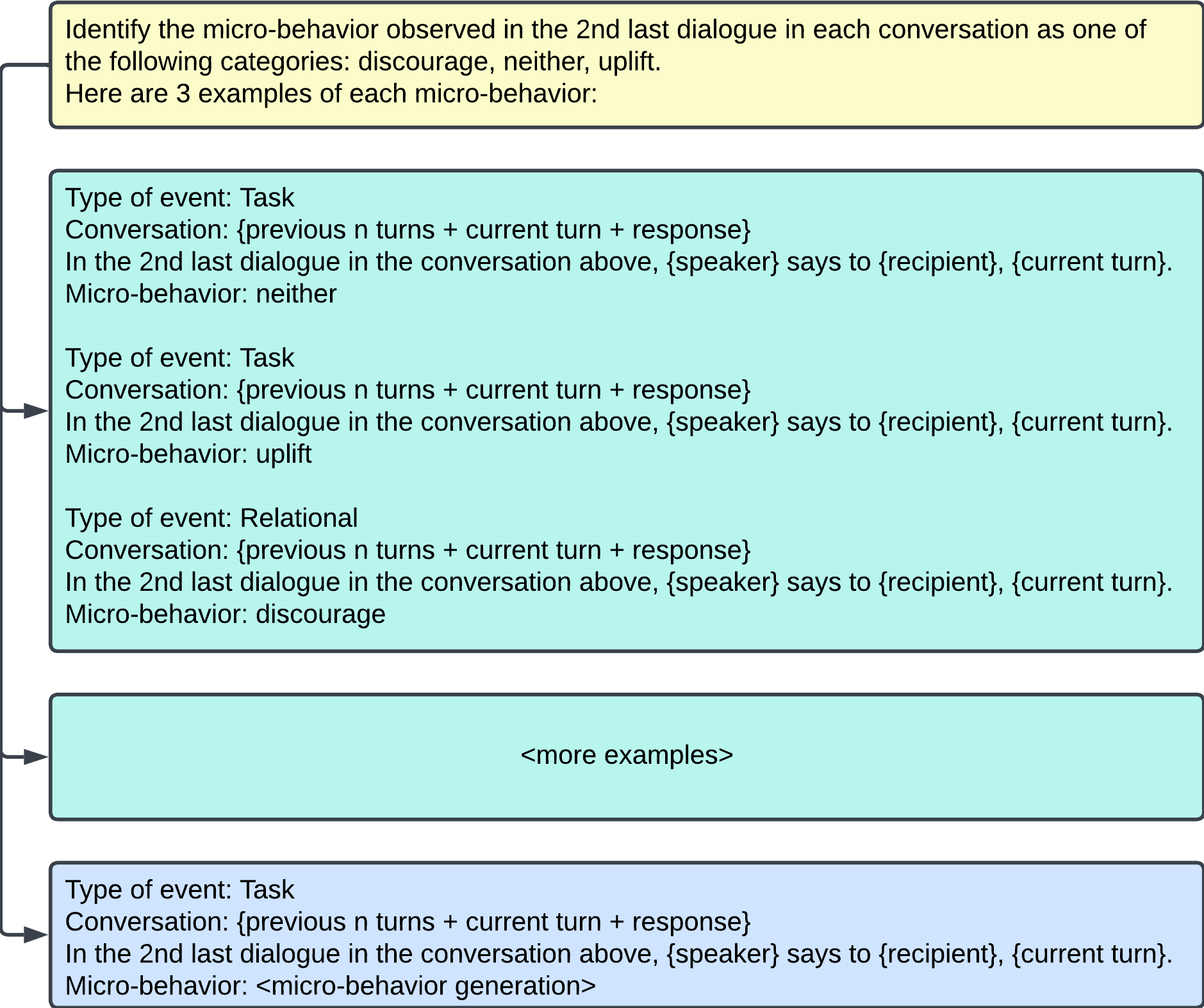}
    \caption{An example of a 3-way classification few-shot learning prompt without the coding definitions for the decoder-only models. The yellow block contains the instructions, followed by few-shot examples in turquoise blocks, and concludes with the test example in the blue block. If applicable, the coding definition is included in the instruction block.}
    \label{fig:prompt}
\end{figure*}

For both of the configurations mentioned above, we initially obtain the zero-shot classification results (i.e., without in-domain fine-tuning), followed by fine-tuning with cross-validation. The fine-tuning pipeline involves holding out 1 team and randomly splitting the input contexts from the remaining 4 teams into training and validation sets in a 3:1 ratio. These models return a single classification label. The parameters are updated using a weighted cross-entropy loss for RoBERTa (i.e., 40\% weightage to the loss on each of the valenced classes, 20\% weightage to the loss on neither class) to address the imbalance between valenced and neutral samples. We also present ablation results by excluding the weighting factors from the cross-entropy loss (i.e., using unweighted/standard cross-entropy loss). In contrast, for DistilBERT, the parameters are updated using unweighted cross-entropy loss, as it is used only for valenced samples. The validation set is used for early stopping in the training loop for both models. This setup is repeated 5 times, with each team serving as the hold-out set once. The results on the hold-out sets of all 5 training runs are aggregated to obtain the micro-behavior classification results for all teams, thus making the comparison with zero-shot pipeline valid. 

We also experiment with generating 2 paraphrases for each conversational turn using the Pegasus paraphraser~\cite{zhang2020pegasuspretrainingextractedgapsentences}, in order to augment the training data threefold. Given the subtle and nuanced nature of micro-behaviors, augmented turns were evaluated by subject matter experts to ensure original conversational semantics were preserved. We report the results while keeping the hold-out set the same as in the standard training process for each cross-validation fold.

\subsection{Decoder-only Text Generation}
\vspace{-5pt}
\label{subsec:DecoderOnly}


The decoder-only causal language modeling pipeline processes the input similarly to how it is handled in experiments involving encoder-only models. The input context includes the current conversational turn, its event type, the previous $k=3,4$ conversational turns, and the response to the current turn, similar to the encoder-only models (Section~\ref{subsec:EncoderOnly}). Additionally, $m$ examples of each micro-behavior are added to the input context to create an in-context learning prompt for the LLMs~\cite{dong2024survey}. We experiment with using $m=3,5,7$ examples of each micro-behavior for the in-context learning prompt. Adding over 4 previous turns or 7 few-shot examples leads to diminishing returns and, in some scenarios, performance degradation, which is why we do not report results with these configurations. We don't report the results when adding only 2 previous turns either, as it is too small a context for the model to appropriately interpret the conversation, confirmed by the poor results on this configuration. Furthermore, we explore incorporating definitions for each micro-behavior, referred to as coding definitions, into the prompt. This complete prompt is then tokenized and passed to the causal language modeling LLM (Figure \ref{fig:decoderonly}). An example prompt is shown in Figure~\ref{fig:prompt}.



We experiment with Llama-3.1~\cite{grattafiori2024llama3herdmodels}, a state-of-the-art causal language modeling LLM developed by Meta. We use the 8 billion parameter Llama-3.1-8B-Instruct model, an instruction fine-tuned version of Llama-3.1, to generate a micro-behavior classification label for the processed conversational turn by learning from in-context examples. The instruction fine-tuned version is better at paying attention to the specific instructions mentioned in the prompt as compared to the base model. During tokenization, the input context is left-truncated to 4096 tokens to provide sufficient space for the instructions and in-context examples. The model’s generation is truncated at 5 tokens and further shortened at the first instance of a newline character. Given the specific instructions that explain the micro-behavior classification task to the model and direct it to the appropriate turn in the conversation context, the model is expected to return one of the 3 micro-behavior classes in the 3-way classification task, or one of the 2 micro-behavior classes in the binary classification task, with the output being directly inferred as is. We quantize these models to 8-bit precision, reducing their size by a factor of 4, which results in faster processing with negligible performance degradation. We consider this problem as a few-shot learning task, similar to previous SLU tasks that were analyzed with LLMs~\cite{mousavi2024llms,zhu2024zero,he2023can}. The few-shot examples for generating the micro-behavior corresponding to a conversational turn are randomly sampled from the conversations of the same team that the turn belongs to. We consider all conversational turns except the in-context learning examples across all teams when reporting the results.

The decoder-only LLM used here has 8 billion parameters, making fine-tuning on our small dataset inefficient, particularly due to the problem of catastrophic forgetting. Nevertheless, we also experimented with fine-tuning using the same cross-validation configuration as the encoder-only models (Section~\ref{subsec:EncoderOnly}) and observed a drastic drop in performance compared to the few-shot prompting scenarios. Therefore, we do not report the fine-tuning results for the decoder-only LLM.

\vspace{-3pt}
\section{Results}
\vspace{-5pt}

The zero-shot macro F1-score with RoBERTa for 3-way micro-behavior classification was 37\% for $k = 3, 4$. Fine-tuning on in-domain data improves performance. Fine-tuning after adding paraphrases to the training data results in further performance improvement with $k=3$, yielding a 41\% macro F1-score. All configurations of the RoBERTa pipeline failed to predict any conversational turn as discouraging, due to the extremely small proportion of discouraging turns in the data. Despite the use of weighted loss, the fine-tuning was still affected by the label imbalance problem, causing the model to favor the majority 'neither' class. The zero-shot macro F1-score for binary classification using DistilBERT was 44\% for $k = 3, 4$. Fine-tuning without paraphrase augmentation, with $k=3$, resulted in the highest improvement, reaching 57\%. Unlike in the 3-way classification with RoBERTa, augmenting the data with paraphrases leads to a lower macro F1-score in binary classification with DistilBERT.

\begin{table}[t]\scriptsize
    \centering
    \setlength{\arrayrulewidth}{.2mm}
    \setlength{\tabcolsep}{4pt}
    \renewcommand{\arraystretch}{1.2}
\begin{tabular}{|l|l|l|l|l|}
    \hline
\textbf{\# previous} & \multirow{2}{*}{\textbf{Zero-shot}} & {\textbf{Standard}} & {\textbf{Weighted}} & \textbf{Paraphrased}  \\
\textbf{turns ($k$)} & & {\textbf{fine-tuning}} & {\textbf{fine-tuning}} & \textbf{fine-tuning}  \\
    \hline
\textbf{3}     & 37 & 36 & 39 & 41 \\  \hline
\textbf{4}     & 37 & 37 & 39 & 31 \\    \hline
\end{tabular}
\caption{Macro F1-scores (\%) for 3-way micro-behavior classification using RoBERTa evaluated under zero-shot classification, standard fine-tuning (unweighted cross-entropy loss), weighted fine-tuning (weighted cross-entropy loss), and weighted fine-tuning with paraphrase-augmented data.}
\vspace{-20pt}
\label{table:roberta_3way}
\end{table}


\begin{table}\scriptsize
    \centering
\setlength{\arrayrulewidth}{.2mm}
    \setlength{\tabcolsep}{4pt}
    \renewcommand{\arraystretch}{1.2}
\begin{tabular}{|l|l|l|l|}
    \hline
\textbf{\# previous} & \multirow{2}{*}{\textbf{Zero-shot}} & \multirow{2}{*}{\textbf{Fine-tuning}} &   \textbf{Paraphrased}  \\
\textbf{turns ($k$)} & & & \textbf{fine-tuning}  \\
        \hline
\textbf{3}       & 44 & 57 & 40  \\  \hline
\textbf{4}    & 44 & 49 & 33 \\    \hline
\end{tabular}
\caption{Macro F1-scores (\%) for binary micro-behavior classification using DistilBERT evaluated under zero-shot classification, fine-tuning, and fine-tuning with paraphrase-augmented data.}
\vspace{-20pt}
\label{table:distilbert_2way}
\end{table}

\begin{figure}[t]
    \centering
    \begin{minipage}[c]{0.49\textwidth}
    \centering
    \includegraphics[width=0.7\linewidth]{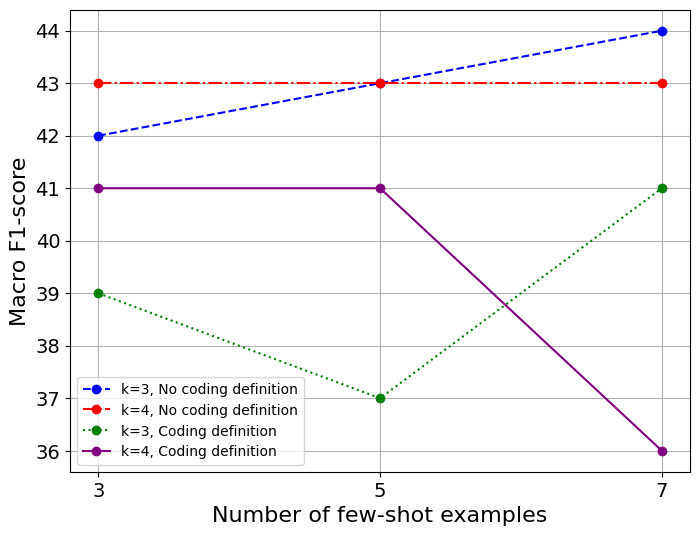}\\
    \vspace{-5pt}
    {{\it (a) 3-way classification}}
    \end{minipage}
    \begin{minipage}[c]{0.49\textwidth}
    \centering
    \includegraphics[width=0.7\linewidth]{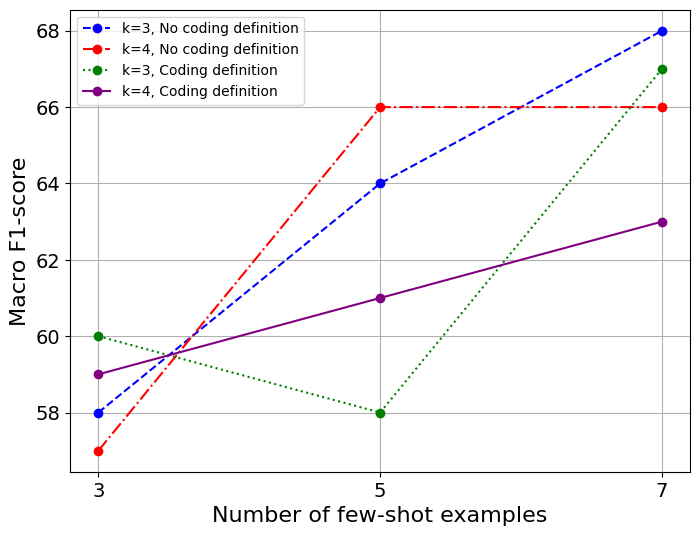}\\
    \vspace{-5pt}
    {{\it (b) Binary classification}}
    \end{minipage}
    \vspace{-8pt}
    \caption{Macro F1-scores for decoder-only Llama-3.1 on 3-way and 2-way micro-behavior classification, respectively, using input with $k=3,4$ previous turns, including or excluding the micro-behavior coding definition.}
    \label{fig:results_llama}
    \vspace{-10pt}
\end{figure}

Using Llama-3.1 (Figure~\ref{fig:results_llama}), majority of the experiments show an overall performance improvement as the number of in-context learning examples increases from 3 to 7. The best Llama-3.1 configurations demonstrate superior performance compared to the best performing RoBERTa (Table~\ref{table:roberta_3way}) and DistilBERT (Table~\ref{table:distilbert_2way}) configurations in both the 3-way (i.e., 44\% vs 41\%) and the binary (i.e., 68\% vs 57\%) micro-behavior classification tasks. Llama-3.1 further demonstrated significantly better recall in detecting uplifting and discouraging micro-behaviors compared to the encoder-only models (Table~\ref{table:results_precision-recall}). This advantage may stem from their ability to move beyond aggregated information across tokens, which is typical in encoder-only models, and instead predict tokens sequentially assigning higher probabilities to distinctive words or phrases characteristic of underrepresented classes. Including the coding definition of micro-behaviors frequently results in minor performance degradation. A potential reason might be that the coding definition was too abstract and the model was not able to fully reason and apply that definition to interpret the task at-hand.


\begin{table}[t]\scriptsize
    \centering
\setlength{\arrayrulewidth}{.2mm}
    \setlength{\tabcolsep}{4pt}
    \renewcommand{\arraystretch}{1.5}
\begin{tabular}{|l|l|l|l|c|}
    \hline
    &   \multicolumn{2}{c|}{\textbf{Llama-3.1}} 
        &   \multicolumn{2}{c|}{\textbf{RoBERTa}}     \\  \cline{2-5}
\textbf{}
    &   \textbf{Uplift} 
        &   \textbf{Discourage} 
    &   \textbf{Uplift} 
        &   \textbf{Discourage}         \\  \hline
\textbf{Recall}       & 55 & 28 & 33 & 1          \\  \hline
\textbf{Precision}    & 30 & 15 & 42 & 23         \\    \hline
\end{tabular}
\caption{Precision and recall for uplifting and discouraging micro-behaviors in the best RoBERTa configuration and the best Llama-3.1 configuration.}
\label{table:results_precision-recall}
\vspace{-20pt}
\end{table}

\vspace{-3pt}
\section{Conclusion}
\vspace{-5pt}
We explored the feasibility of LLMs in detecting micro-behaviors in team conversations during space missions, using conversation transcripts alone. While fine-tuning encoder-only models improved the detection of uplifting and neutral conversations as compared to zero-shot classification, they failed to identify discouraging conversations, possibly due to label imbalance in the training data. Additionally, our observations indicate that data augmentation through paraphrasing does not always lead to performance improvement and may even degrade performance if the original dataset is too small, as observed during the fine-tuning of DistilBERT. In contrast, the decoder-only Llama-3.1 model demonstrated superior performance, detecting both uplifting and discouraging classes with as few as 7 in-context learning examples. Including 3 previous conversational turns in the input context seems sufficient for these models to understand the context surrounding a conversational turn. Future work will assess the generalizability of these findings beyond space applications and will incorporate multimodal signals, such as prosody and nonverbal cues.

\bibliographystyle{IEEEtran}
\bibliography{mybib}

\end{document}